\documentclass{article}
\usepackage{arxiv}

\usepackage{cite}
\usepackage{amsmath,amssymb,amsfonts}
\usepackage{algorithmic}
\usepackage{multirow}
\usepackage{graphicx}
\usepackage{textcomp}

\usepackage{verbatim} 
\usepackage{soul} 
\usepackage{wasysym} 
\usepackage{todonotes}
\usepackage[T1]{fontenc}
\usepackage[utf8]{inputenc}

\title{OSN Dashboard Tool for Sentiment Analysis}

\author{
Andreas Kilde Lien,
Lars Martin Randem, Hans Petter Fauchald Taralrud, Maryam Edalati 
\\
 \textit{Department of Computer Science} \\
    \textit{Norwegian University of Science and Technology}\\
    Gjøvik, Norway \\
    \{andrklie, larsmran, hptaralr, maryame\}@stud.ntnu.no
}

\begin{document}
\maketitle

\begin{abstract}
The amount of opinionated data on the internet is rapidly increasing. More and more people are sharing their ideas and opinions in reviews, discussion forums, microblogs and general social media. As opinions are central in all human activities, sentiment analysis has been applied to gain insights in this type of data. There are proposed several approaches for sentiment classification. The major drawback is the lack of standardized solutions for classification and high-level visualization. In this study, a sentiment analyzer dashboard for online social networking analysis is proposed. This, to enable people gaining insights in topics interesting to them. The tool allows users to run the desired sentiment analysis algorithm in the dashboard. In addition to providing several visualization types, the dashboard facilitates raw data results from the sentiment classification which can be downloaded for further analysis.

\end{abstract}


\section{Introduction}
\label{sec:intro}

Sentiment analysis or opinion mining is the computational study of people’s opinions, sentiments, emotions, appraisals, and attitudes towards entities such as products \cite{fang2015sentiment,bibi2022novel,zhao2017weakly}, services \cite{10.1145,sadriu2022automated,edalati2021potential} \cite{feldman2011stock}, organizations, individuals \cite{batra2021evaluating}, education \cite{kastrati2021sentiment,sandra2022university,kastrati2020weakly}, issues \cite{nezhad2022twitter,shariq2020cross}, movies \cite{ali2019sentiment,yasen2019movies}, events \cite{Fu:2016,imran2020cross,electronics10101133}, topics \cite{stamm2022does}, and their attributes \cite{liu_2015,dalipi2017analysis}. The main idea with sentiment analysis is to detect polarity in textual data and predict the polarity as either positive, negative or neutral. With the abundance of opinionated data on the internet, this Natural language processing (NLP) technique can be applied to obtain perspective in various topics. Sentiment analysis is applicable when performing decision making tasks. If the uncertainty arises when you are looking to buy a new product, customers will be influenced by product reviews and discussion forums to chose the right product \cite{bibi2022novel}. When conducting business, sentiment analysis can provide the organization important insights in understanding the customers’ feelings towards a product. Since online social media platforms allow users across the world to share ideas, thoughts and feelings, opinion mining is a powerful tool to observe cultural differences because people reason differently \cite{JohnsonLaird2006AreTC}. However, recent research shows that countries with similar cultural features acted differently dealing with the coronavirus situation \cite{imran2020cross}.
Sentiment analysis tasks can be carried out at different levels of granularity \cite{kastrati2021sentiment}. The first level is at document level. This analysis is predicting the polarity of sentiments found in the entire document. Second is sentence level sentiment analysis. This refers to classifying the polarity of each sentence. The last level of granularity is the aspect level. Aspect level analysis entails predicting the polarity of different aspect words in a sentence.

Dashboards provide visual representations of data \cite{doi:10.1177/1094670509344213}. One of the key features with dashboards, is that they present at-a-glance information. The information on a data dashboard should be easy to understand and look at, rather than complex visualizations that takes a lot of time to process. For example, a car dashboard should be easy to interpret. If it demands too much time to look at, the driver will eventually crash. Dashboards are heavily used in business as it facilitates high-level insights of real time data to monitor the business performance \cite{AzureGroup2020}. In addition, it can enhance the decision making of the business. The set of visualization varies from use case to use case, but common visualization types are line charts, bar charts, pie charts and scatter plots. However, certain types of visualization have pre-attentive features where humans can interpret them before paying attention. These kinds of visualization include comparing the length of elements (bar chart) and their positions in two-dimensional space (line charts) \cite{NielsenNormanGroup2020}.

In this article, the main objective is to create a sentiment analyzer dashboard for online social networking analysis. The study is focusing on collecting social media posts from Twitter, as it is one of the most widely used social platforms, having over 365 million users \cite{BrianDean2022}. In addition, the publicly available Twitter API allows for easy fetching of posts. The OSN dashboard will fetch Twitter posts based on given input criteria from the user and perform document sentiment analysis on them. The dashboard will provide an overview of the sentiment polarity in those tweets and the ability to identify trends on people across the world towards a topic of interest.

The contributions of this research work are the following:
\begin{itemize}
    \item A sentiment analyzer dashboard for Twitter data.
    \item Perform sentiment analysis on various topics.
    \item Apply different sentiment analysis algorithm on Twitter data. 
    \item Monitoring interesting trends across the world.
    \item Accessing raw data for further analysis.
\end{itemize}

The rest of the article is structured as follow. Section \ref{sec:relatedWork} entails related work regarding sentiment analysis and dashboards. The method, including architecture, design and implementation, is presented in section \ref{sec:methodology}. In section \ref{sec:results}, we present results and discussions, followed by conclusion and future work in section \ref{sec:con}.

\section{Related Work}
\label{sec:relatedWork}

\subsection{Sentiment Analysis}
\label{sec:SA}

Sentiment analysis has become a popular research field in natural language processing (NLP) \cite{9237640}. Particularly, performing sentiment polarity assessment on Twitter data has been applied in various application domains including forecasting stock prices \cite{8806182,8621884,8537242}, students’ feedback \cite{kastrati2020weakly} and predicting president elections \cite{8399007}. There is a great amount of literature on the subject. For example, \cite{kastrati2021sentiment} provide a systematic mapping study of reviewing literature on sentiment analysis in the context of the education domain. The results show 92 relevant studies conducted between 2015 and 2020. This supports that the field is extensively researched and rapidly growing.
There are proposed several approaches to automatic sentiment categorization. Methods can be lexicon-based \cite{Maite2011Lexicon} and dictionary-based \cite{9277094}. However, recent literature show a transfer from pure NLP techniques to machine learning and deep learning approaches \cite{Zhang2018Deep}. The major drawback with the methods for sentiment analysis, is the lack of standardized solutions. Current solutions are programming language dependent and perform only certain tasks \cite{kastrati2021sentiment}. Some existing sentiment analysis algorithms are TextBlob, VADER, Flair and Stanza. Table \ref{tab:performance} shows an overview of their approach and the accuracy achieved by them on the Sentiment140 dataset.

\begin{table}[!htb]
    \centering
    \begin{tabular}{c|c|c|c}
    \hline
       Model  & Name & Dataset & Accuracy  \\\hline
         Lexicon & TextBlob & Sentiment140 & 65.06 \\
         Lexicon & Vader & Sentiment140 & 72.29 \\
         LSTM & Flair & Sentiment140 & 94.09 \\
         CNN & Stanza & Sentiment140 & 88.18 \\
         \hline
    \end{tabular}
    \caption{Performance of sentiment analysis algorithms}
    \label{tab:performance}
\end{table}

The authors in \cite{DBLP:journals/corr/abs-1709-02984} address the issue that off-the-shelf sentiment analysis tools are primarily trained on general social media data. Therefore, a classifier trained to support sentiment analysis in developers’ communication channels is proposed to accommodate jargon within technical domains. The model exploited a series of lexicon-based, keyword-based \cite{rajput2016lexicon, puigcerver2017querying}, and semantic-based features \cite{kastrati2016semcon, tsatsaronis2010semanticrank, kastrati2015semcon}. It was trained on a dataset consisting of over 4000 posts on Stack Overflow. With respect to an off-the-shelf sentiment analysis baseline, SentiStrength, results show that the proposed model improves 19\% in precision for negative polarity and a 25\% improvement in recall for the neutral class.

Research conducted by Imran et al. \cite{imran2020cross} propose cross-cultural polarity and emotion detection in the context of COVID-19 tweets. Deep learning models were used to classify positive and negative sentiments and corresponding emotions in tweets. The performance of the trained models provided state of the art accuracy of the sentiment140 dataset.

Due to the fact that many people express feelings in their native languages, literature has attempted to develop approaches for multilingual sentiment analysis \cite{chandio2022attention, batra2021large}. Zhu et al. propose a semi-supervised method based on bootstrapping and a SVM classifier to predict sentiment and polarity classification on microblog data \cite{10.1007/978-1-4614-6880-6_28}. Various combination of features such as the word and part of speech were used to improve the performance. A problem with multilingual sentiment analysis is the lack of lexical resources \cite{fatima2022systematic, Kia2016Multilingual}. This can be solved by utilizing translation systems \cite{ghafoor2021impact} and synthetic data \cite{imran2022impact} which also can affect the validity of the original resource. In that case, the authors in \cite{8628718} propose a concept-level knowledge base for multilingual sentiment analysis which is available in 40 languages.

\subsection{Dashboards}
\label{sec:DB}

Mahadzir et al. \cite{ijtech-2753} developed a sentiment analysis dashboard using real-time Twitter data, with the objective to understand the imbalance between supply and demand in the property industry of Malaysia. Data was extracted by limiting the Twitter posts to a specific keyword. Twitter posts were retrieved in both Malay and English. The authors performed sentence-based and aspect-based classification using a Naıve Bayes classifier to predict the polarity of the overall tweet and the polarity of the aspects within the post. The dashboard provided an overview of the overall sentiment and aspect-based analysis, real-time Twitter monitoring statistics and the detail of tweets based on their features and polarity. There are some limitations with this study. First, it is domain specific and lets users only get an overview of the property industry in Malaysia. Second, only 745 tweets was collected related to the domain. Maybe utilizing other social media platforms would resulted in more data. Lastly, the data was classified using one machine learning algorithm. By supporting the users to choose the desired sentiment analysis approach to be used in the dashboard, the results may change.

Most existing sentiment analysis dashboards are highly based on customer reviews. Different organizations use these dashboards to retrieve feedback from customers related to specific products and services. The solutions mine opinions from websites such as Amazon Reviews \cite{7919584}, Google Reviews and social media including YouTube, Twitter, Twitch and TikTok. Brandwatch 1 and Repustate 2 are examples of such customer intelligence platforms. The former is using a hybrid approach of manual and automated NLP techniques when assessing sentiments. Their approach is structured in three steps: In the first step, a sentence is processed trough a knowledge based rule and is classified as positive, neutral or negative. The rules are based on common language and does not deal with domain specific language. The second step deals with sentences that the knowledge based system is unable to classify. Here, the sentence is processed through a machine learning classifier that is taught to understand technical jargon. Lastly, the third step enables users to customize rules that are domain specific which will enhance the accuracy of the model. Repustate uses an extensive multilingual sentiment analysis approach which classifies sentiments in 23 languages. This can be helpful for international brands to perceive different opinions across regions and cultures.

\section{Methodology}
\label{sec:methodology}

This section entails the architecture aspect of the application and the funcational requirements for the tool. 

\subsection{Architecture}
\label{sec:archi}

\begin{figure}[ht]
    \centering
    \includegraphics[width=15cm]{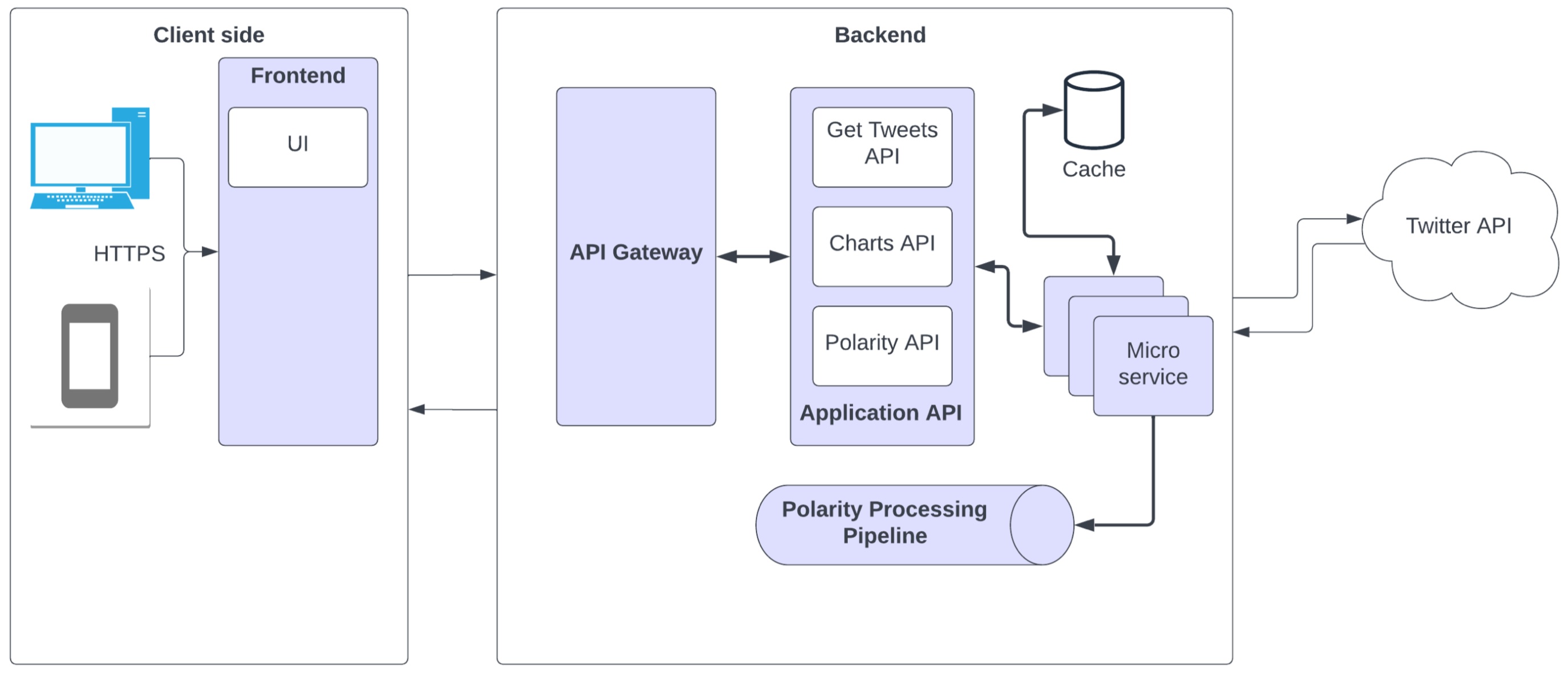}
    \caption{Evolution of energy prices, 2020-2021.}
    \label{fig1}
\end{figure}

The high-level architecture for the dashboard is made up of a microservices architecture, as shown in Figure \ref{fig1}. The main separation in the architecture is between the client side and backend. The client side is for the content that is displayed on the end-devices such as PC and smartphone while the backend is where the data is gathered and processed. To give a more detailed description of each component in the architecture: On the client side is a frontend service represented as a web application. In the back-end there is an "API Gateway" that handles all the tasks involved in accepting and processing up to hundreds of thousands concurrent API calls, that are presented in the "Application API" service. The API calls gather the data from the respective microservice that handles the part of the process. For the most demanding tasks, there will be a caching service, for example collecting a lot of tweets in real-time. Connected to the polarity microservice is a "Polarity Processing Pipeline" to handle big data in real-time with regards to social media posts. The Twitter API is added as a 3rd party service connected to the backend.

A microservices architecture offers several benefits. Since the functionality is decoupled into separate services, it becomes possible to deploy and make changes to each service independently from the rest of the application. Whereas in a monolithic application, adding new features requires the entire application to be redeployed, and changing one feature might also result in unwanted side effects in other parts of the application \cite{Google2021}. For example, with a microservices architecture, we are able to easily change our microservice that is responsible for the sentiment analysis.

Another benefit of microservices architecture, is that it is easier to maintain each microservice. A monolithic application can become quite complex as it grows, with many different features that depends on each other. With a microservices architecture, each microservice only has one responsibility \cite{Google2021}.

Having an application split into separate microservices also allows each microservice to be written in different programming languages. This allows the programmers to choose the language that is best suited for a microservice’s task \cite{Google2021}.

There are however some drawbacks that needs to be taken into account when using a microservices architecture. Microservices increases the complexity of the application, as it introduces more points of failure, since microservices needs to communicate with each other. The application needs error handling to take care of cases where a microservice is unavailable. If a microservice fails, it is important that data between each microservice stays consistent \cite{Google2021}.

An application with a microservices architecture might also perform slower due to network latency when microservices communicate with each other, while a monolithic application can call functions directly \cite{Google2021}.

\subsection{Functional requirements}
\label{sec:FR}

As introduced in Section 1, the tool will fetch Twitter posts and provide a sentiment analysis overview on the given different topics. To be able to reach the end goals, we formulated functional requirements which describe the functionality of the system:

\begin{itemize}
    \item The system must allow users to search on Twitter posts based on keywords, hashtags and usernames.
    \item The system must allow users to perform a extended search on Twitter posts including language, time frame, origin and the number of posts.
    \item The system must allow users to perform sentiment analysis on the Twitter posts.
    \item The system must allow users to choose the desired sentiment analysis algorithm to use in the dashboard.
    \item The system must provide appropriate data visualization such as plots, word cloud and map.
    \item The system must allow users to monitor the raw data provided by the dashboard.
    \item The system must allow users to download the raw data in a \textit{.csv} format.
\end{itemize}

The following non-functional requirements are also created:

\begin{itemize}
    \item The user must be able to perform a simple search within max two steps.
    \item The application should handle 100 requests simultaneously when requesting less than 1000 Tweets.
    \item When requesting x Tweets each page must load within $x \times 0.05$ second(s).
    \item The system must meet Web Content Accessibility Guidelines WCAG 2.1.

\end{itemize}

\subsection{Implementation}
\label{sec:imp}

The backend will consist of a Flask\footnote{https://flask.palletsprojects.com/en/2.1.x/} API framework and Redis\footnote{https://redis.io/} for database cache. Redis will be connected to API calls that fetches data from the Twitter API. This is done to make the API calls faster and less fragile for DDOS attacks. The cache time will be set to 60 seconds, because the user expects the latest Tweets to be returned.

\begin{figure}[ht]
    \centering
    \includegraphics[width=15cm]{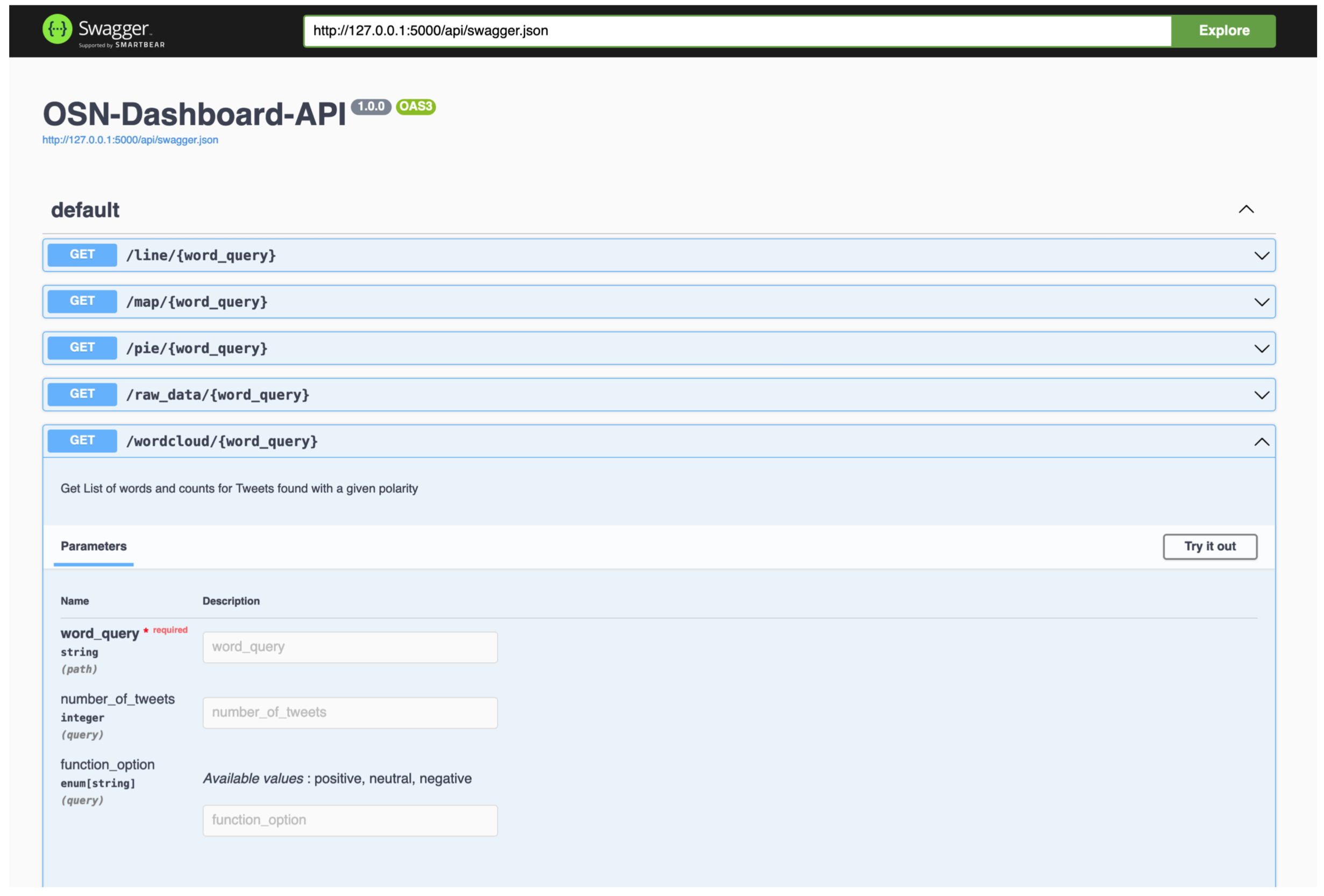}
    \caption{Screenshot of the Swagger API documentation.}
    \label{fig2}
\end{figure}

The API will be a REST API that have multiple API calls. Each call will be independent to each module in the frontend (pie chart, raw table data, etc.). The documentation will be provided with Swagger\footnote{https://swagger.io/} that also functions like a user interface for the API calls, like shown in Figure \ref{fig2}. Each call will be made with focus on minimising bigger calls to the Twitter API, because that is demanding and time consuming.

The frontend is built using the web framework React\footnote{https://reactjs.org/}.

Table \ref{tab:SecurityRequirements} summarizes the major security concerns for the tool. 
\begin{table}[!htb]
\caption{Security Requirements}
\label{tab:SecurityRequirements}
\begin{tabular}{|p{1cm}|p{8cm}|p{4.5cm}|p{1cm}|}
\hline
Type &
  Requirement Description &
  Comments &
  Priority \\ \hline
\multirow{3}{*}{\rotatebox{90}{Authentication}} &
The system shall have
authentication measures
at all the entry
points or inbound
network connection. &
  To avoid unauthorized access &
  1 \\ \cline{2-4} 
 &
  The system shall support
authentication
based on a API token
with the back-end. &
  Improving the security
using a unique token. &
  1 \\ \cline{2-4} 
 &
  The system shall only
allow incoming network
requests from
within a network. &
  To avoid unauthorized
access. &
  2 \\ \hline
\multirow{2}{*}{\rotatebox{90}{Availability}} &
  There has to be a hard
limit for how many
Tweets to fetch. &
  To not create a huge
overhead. &
  3 \\ [0.2cm]\cline{2-4} 
 &
  The system shall apply
caching of Twitter
data &
  Help minimizing the
impact of potential
system failures. &
  3 \\ [1cm]\hline 
\rotatebox{90}{Auditing} &
The system shall keep
historical records (logging)
of events and processes executed in
or by the application. &
  Define more specific security
loggings to allow
recreating a clear picture
of security events. &
  1 \\ \hline
\multirow{3}{*}{\rotatebox{90}{Authorization}} &
  The user token shall
possess privileges
within the application
to perform their
activities. However,
the privileges must be
limited. &
  Avoid an unauthorized
user execute activities
as another user. &
  1 \\ \cline{2-4} 
 &
  The system shall
ensure system level
accounts have limited
privileges. &
  Help avoiding attackers
escalate user’s accounts
to access administrator’s
features. &
  1 \\ \cline{2-4} 
 &
  The system shall ensure
the Twitter token
is performed using parameterized
store procedures
to allow all access
to be revoked. &
  Apply security principles. &
  1 \\ \hline
\end{tabular}

\end{table}

\begin{figure}[ht]
    \centering
    \includegraphics[width=15cm]{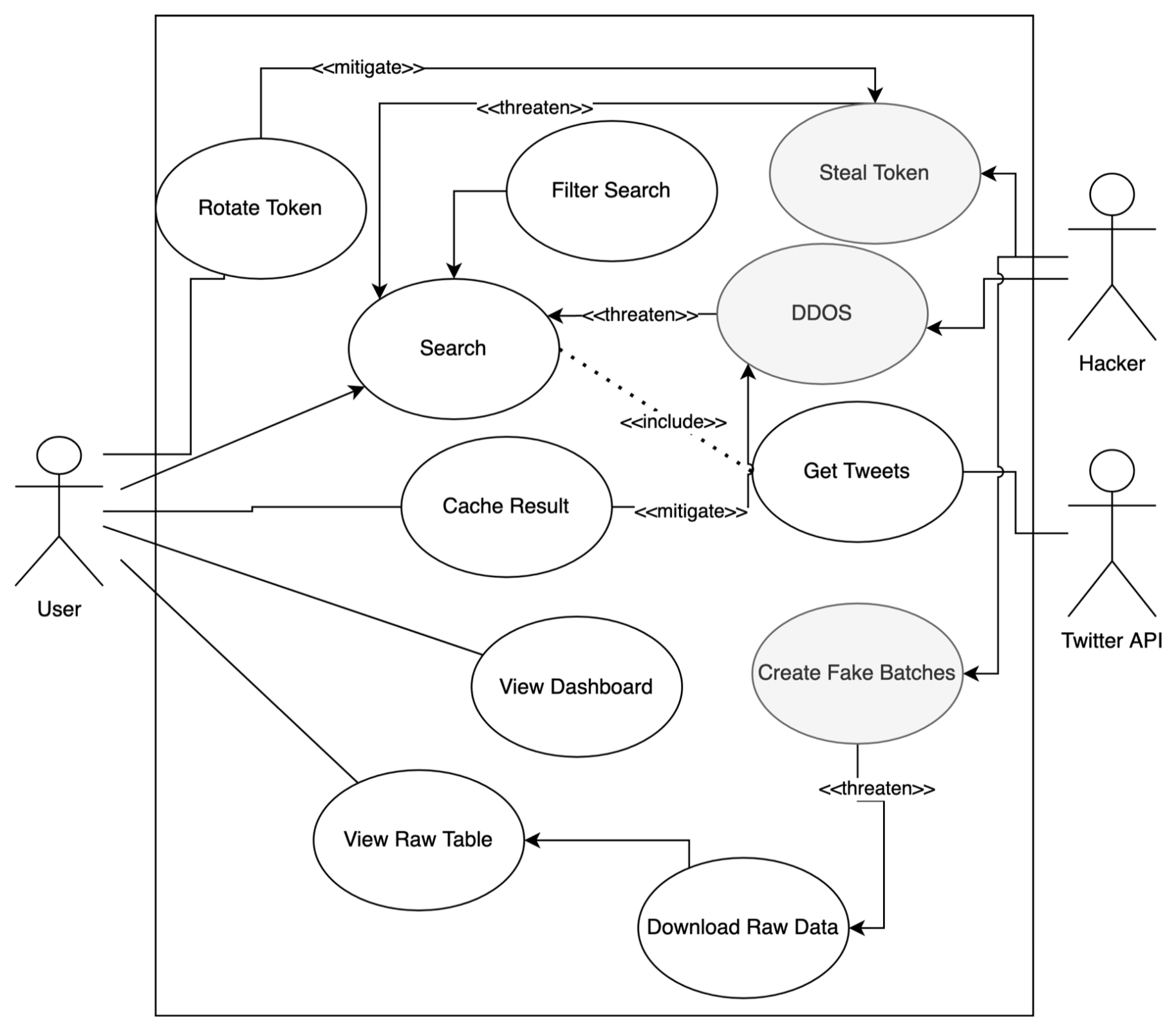}
    \caption{The figure illustrates the missuse/abuse case. License: Andreas Kilde Lien, CC-BY-4.0.}
    \label{fig3}
\end{figure}

The research work in \ref{fig3} illustrates how a hacker can exploit the system. In this Section there is created an abuse case description for one misuse case. This approach allows the Test Analyst to create test cases for Security Requirements.

\begin{figure}[!htb]
    \centering
    \includegraphics[width=\textwidth]{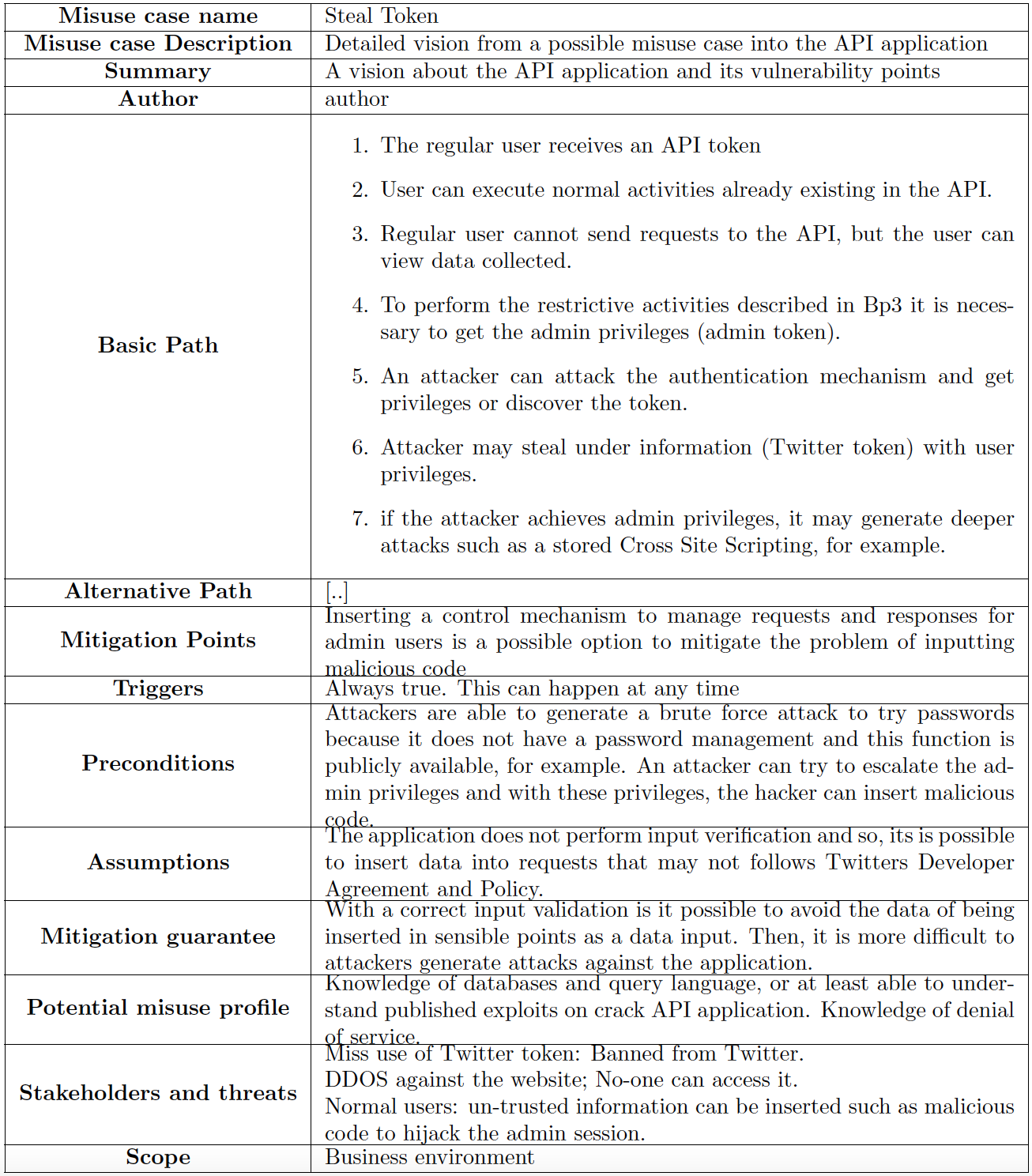}
    \caption{Misuse cases}
    \label{fig. misusecases}
\end{figure}

\subsection{Design principles}
The following design principles are considered in this study. 

\textbf{Least privilege:} When utilizing the API, different users should have different privileges. By following the least privilege principle that minimize the occurrence of unintentional, unwanted, or improper uses of privileges.

\textbf{Minimizing attack surface area:} The minimizing attack surface principle will be used to minimize the entry points to the application. Since the Token from Twitter has a limit in number of fetch data, it is required to limit access for only authorized users and the application can only be used within an institution's network.

\textbf{Economy of mechanism:} This principle will be covered by implementing simpler and smaller functions that are easy to maintain.

\textbf{Fail securely:} It is important that the application do not crash unexpectedly. Therefore, will the fail securely principle by handling errors, such as timeouts and fault inputs.

\textbf{Do NOT trust:} This principle will be implemented by restricting user’s access, and validate user input.

\section{Results \& Discussions}
\label{sec:results}

\begin{figure}[ht]
    \centering
    \includegraphics[width=15cm]{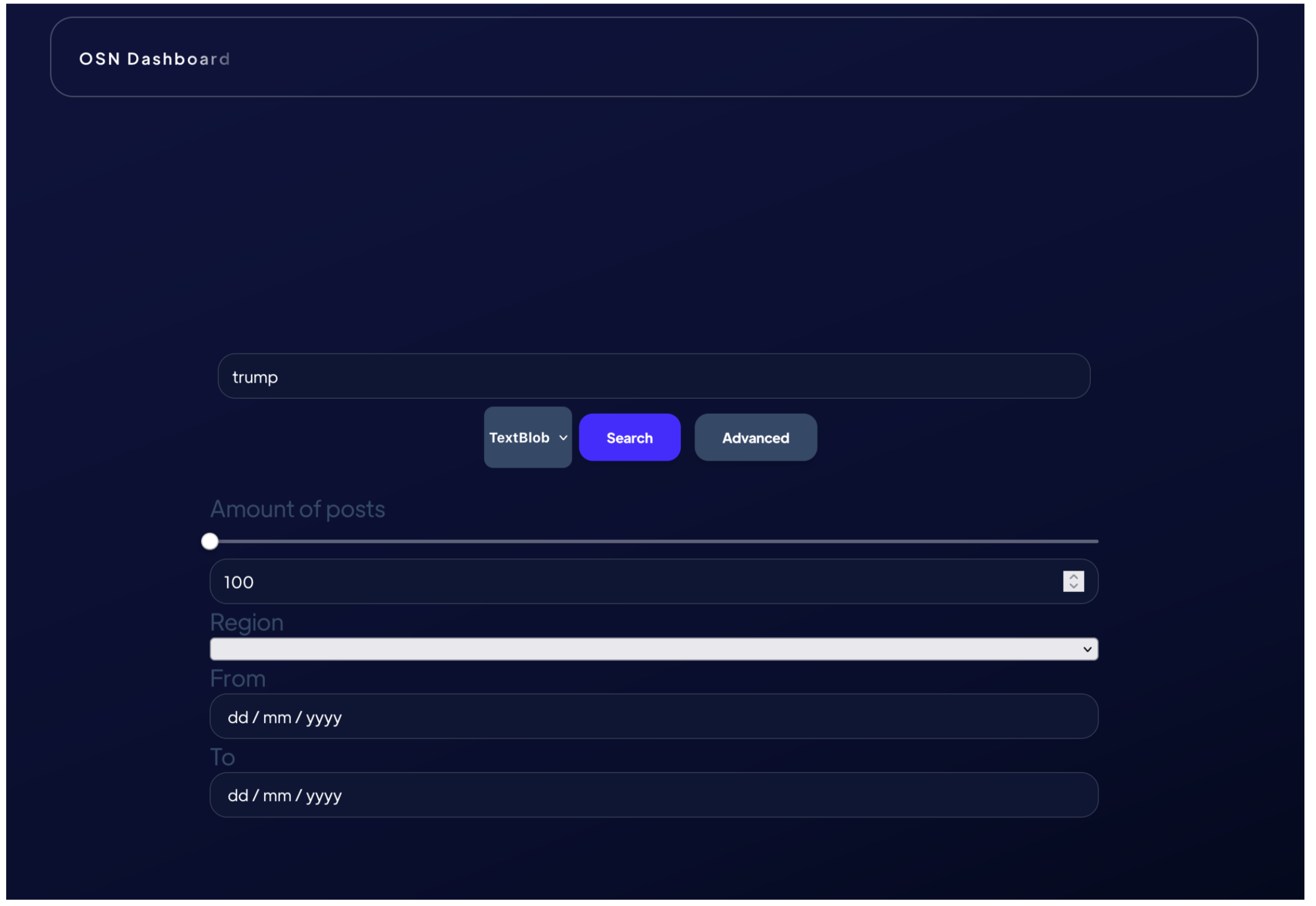}
    \caption{search page}
    \label{fig4}
\end{figure}

Following the design principles described in Section 3.4, the final product is created. Figure \ref{fig4} illustrates the search page. Here, users can search on tweets based on keywords, username or hashtags. In addition the user can choose a preferable sentiment algorithm to use in the dashboard. The users can also extend the search by applying more criteria in the advanced search page. 

\begin{figure}[ht]
    \centering
    \includegraphics[width=15cm]{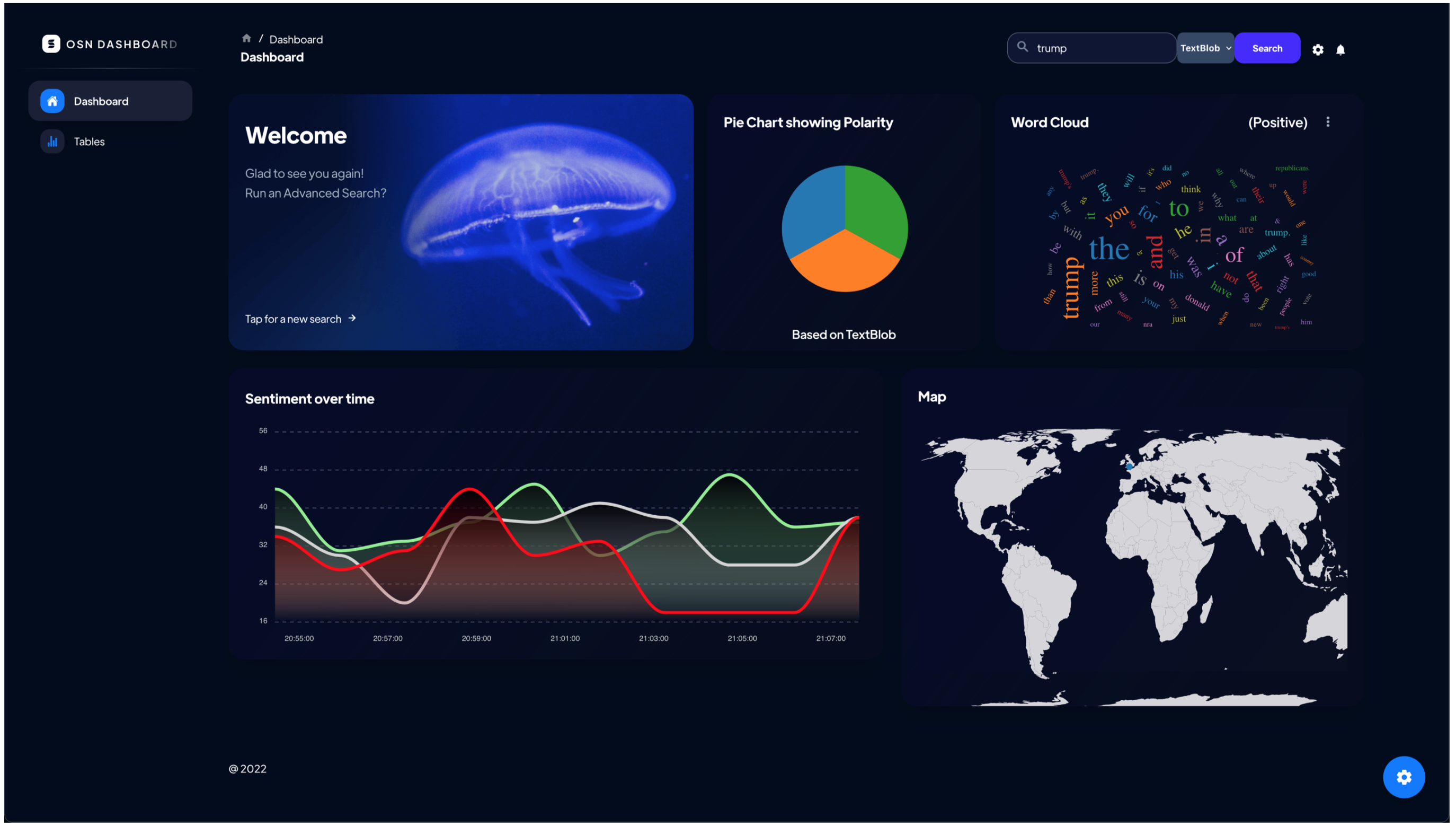}
    \caption{The dashboard page}
    \label{fig5}
\end{figure}

As a search is performed, the user will arrive the dashboard page (Figure \ref{fig5}), where the user can visualize pie chart, line chart, tag cloud and map. In the top of the screen a new search can be performed.

\begin{figure}[ht]
    \centering
    \includegraphics[width=15cm]{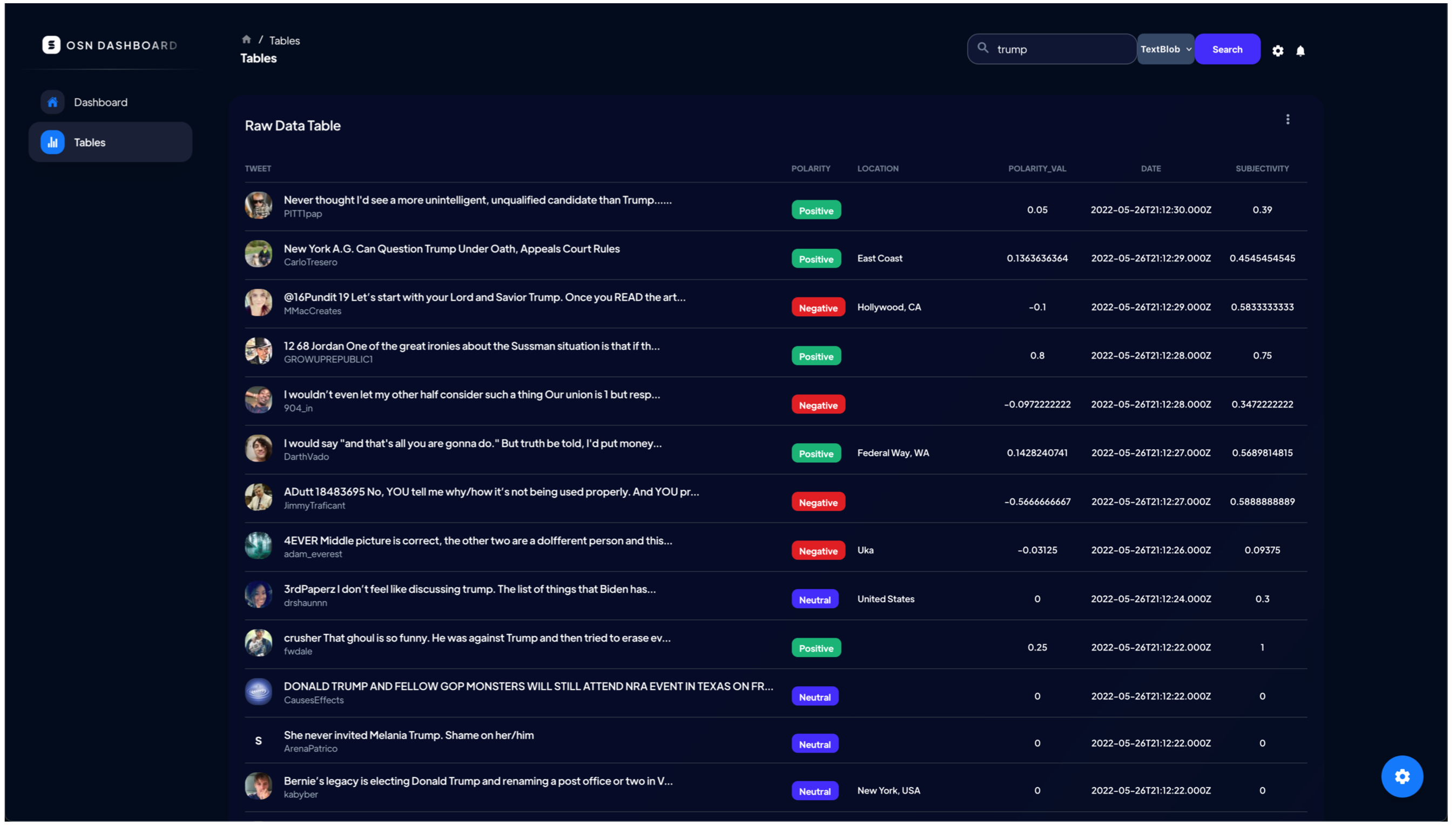}
    \caption{The dashboard page}
    \label{fig6}
\end{figure}

By clicking the "Table" button in the side navigation bar, direct the users to the raw data page depicted in Figure \ref{fig6}.

The OSN dashboard is also a responsive website. It can be scaled such that it works fine on smaller devices such as mobile and tablets. The dashboard is using a dark blue color scheme. The reason behind the choice of color is based on how colors arouse emotions in users. Based on research in color psychology, the color blue is associated with trust and stability \cite{10635_99059}.

To assess the novelty of the OSN dashboard, two models named Level of Inventiveness and Norwegian Research Council (NRC) Scale are used.

\begin{figure}[ht]
    \centering
    \includegraphics[width=12cm]{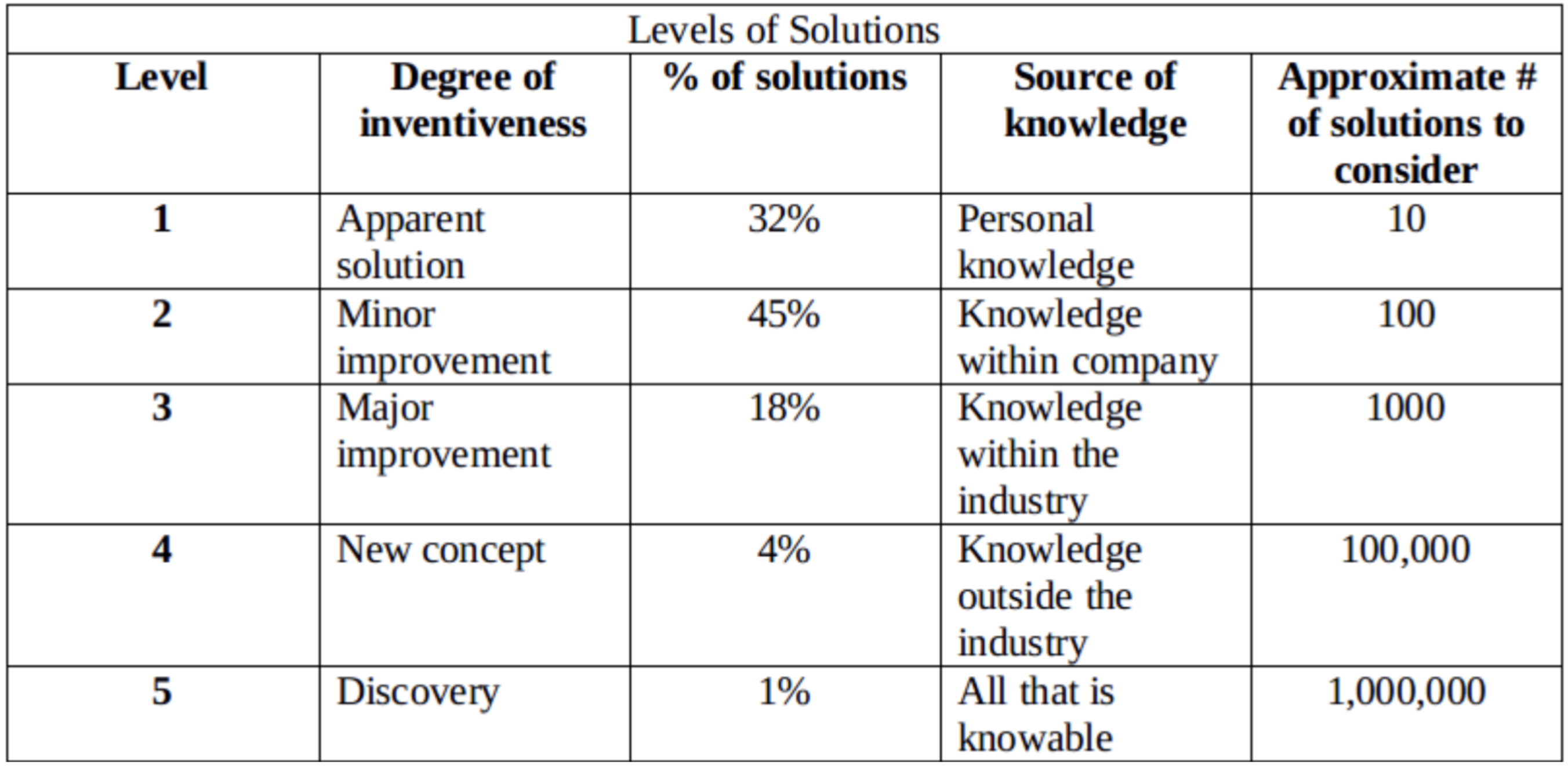}
    \caption{Level of Inventiveness scale}
    \label{fig7}
\end{figure}

\textbf{Level of Inventiveness:}
With respect to the Level of Inventiveness scale depicted in Figure \ref{fig7} \cite{Softqube2022}, we believe the study is at level three. The major improvements are that the OSN dashboard is a more conventional way to perform sentiment analysis on online social media data. It is possible to search on various topics of interest and receive high-level insights. With the use of visualization such as map, the users will perceive how different regions and countries express different opinions about the same topic. Another improvement is that researchers can choose the desired sentiment analysis algorithm to use in the dashboard and compare their results. In addition, integrating sentiment analysis algorithms which support multilingual posts will provide more and better results on tweets around the world. All these above mentioned improvements are knowledge within the industry, but the solution differ from the industry competitors in the way that search on topics do not need to be domain specific and it provides an intuitive way to perform sentiment analysis.

\begin{figure}[ht]
    \centering
    \includegraphics[width=12cm]{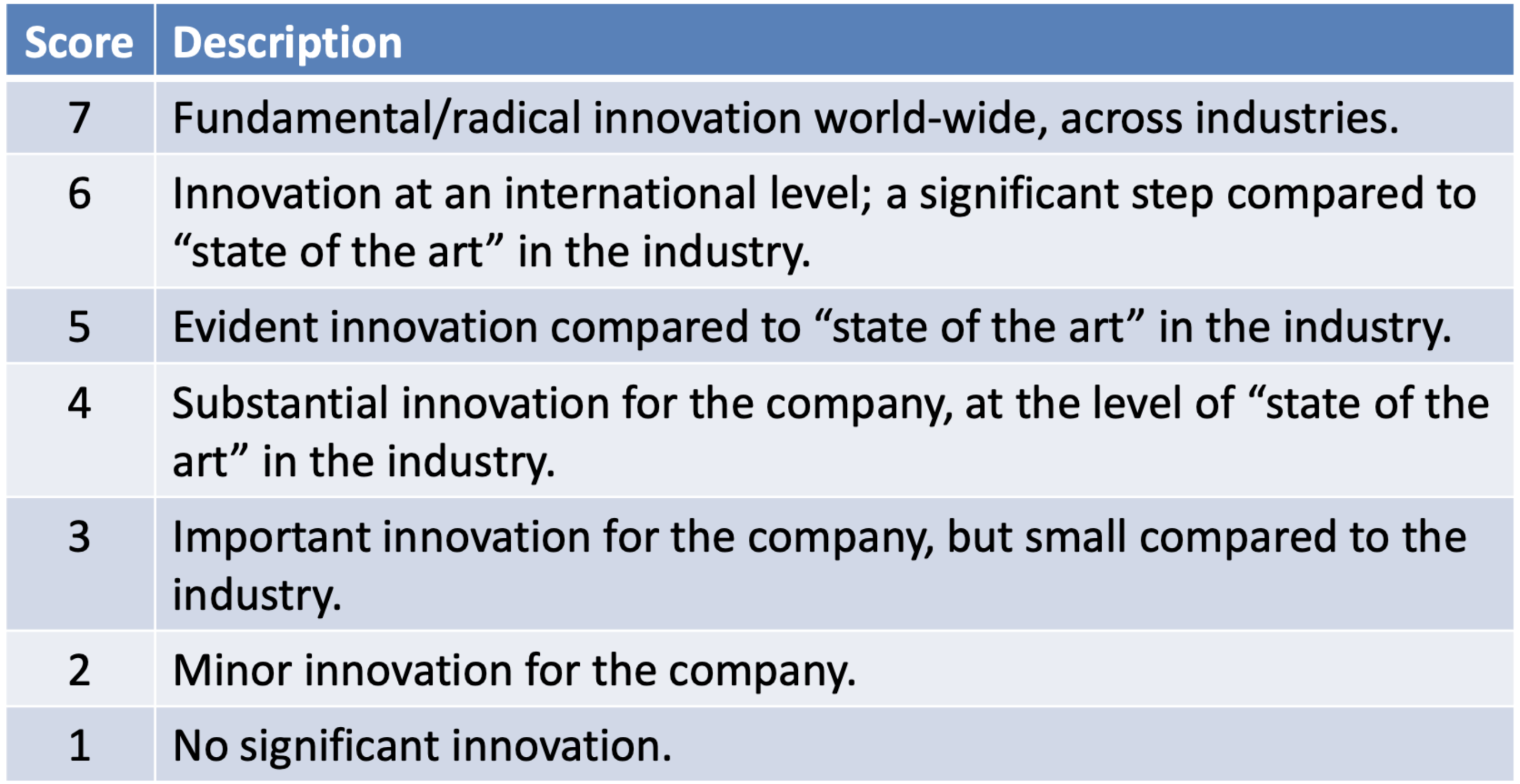}
    \caption{NRC scale}
    \label{fig8}
\end{figure}

\textbf{NRC Scale:}
In regards to the NRC scale shown in Figure \ref{fig8} \cite{Council2020}, the OSN dashboard is given a score of four. Recall the improvements mentioned in Section 4.2.1, the tool is a substantial innovation as it provides a new combination of knowledge within the industry. A concern with assessing the OSN dashboard is how to evaluate if it is at the level with the state-of-the-art in the industry. The dashboard consists of aspects which are not directly countable in any measure. By utilizing sentiment analysis algorithms, it will naturally perform similarly. However, the assessment of how intuitive the dashboard is requires usability testing and the need for the users to compare the tool with stat-of-the-art sentiment analysis dashboards in the industry.

\subsection{Integration}
\label{sec:integration}

There are several application domains the OSN dashboard is suited for. First and foremost the tool can be integrated to the Twitter application. Here, users can retrieve public opinions regarding news, products, politics etc. Second, the dashboard can have huge impact on any decision making process and can be used with existing frameworks for getting students' opinions in education for instance, such as in \cite{imran2012multimedia, misuraca2020sentiment, imran2014hip}. Another example can be an organization such as Foodora\footnote{https://www.foodora.com} which can integrate the tool and utilize it as a part of their market campaign because the peoples opinions influence how they will conduct the business in the future. Another example of decision making is in the financial domain. The finance world becomes more and more quantitative. However, it has been shown that social media can affect the stock market to a great extent \cite{COSTOLA2021110021}. Therefore, the OSN dashboard can be integrated to the portfolio management tools to collect qualitative data. This data will support any decision towards investments. The study however doesn't incorporate embedding models \cite{kastrati2020wet} and advanced deep learning algorithms for sentiment analysis \cite{zhang2018deep12, sun2019utilizing}. So, the support for these can be provided in the tool in future. 

\subsection{Deployment}
\label{sec:deployment}

The complete application is Dockerized\footnote{https://www.docker.com/} to ship all the applications with all the necessary functionalities as one package with Docker Compose (multi-container).
The Twitter API has both an app rate limit and user rate limit that is controlled around requests per minute. Twitter has an options for a paid "Premium" API that allows many more requests per minute. As long as the application will not have to many users, using the public API should be sufficient.

\section{Conclusion \& Future Work}
\label{sec:con}

The OSN analysis dashboard is developed to enable users gaining insights in various topics. This, to get a perspective of the world. The tool can be utilized for several purposes such as decision making and research. Therefore, the dashboard can be integrated to several application domains including the Twitter website, product websites or portfolio management tools.
In this article, an implementation of a sentiment analyzer dashboard was proposed. Sentiment analysis was performed on Twitter data based on keywords, hashtags or usernames given by the user. Different sentiment analysis algorithms were integrated to perform sentiment categorization. The analysis results are visualized in form of a dashboard to provide at-a-glance information. The data on the dashboard is presented in appropriate plots and charts with the addition of raw data from the sentiment analysis results.

When it comes to future work, there are always room for improvements as not every implementation goal is reached within the given time frame. Currently, the sentiment analysis algorithms integrated are lexicon-based. It would be interesting to add algorithms which are machine learning-based to compare the differences and train them on domain-specific topics using ontology \cite{kastrati2015improved, nirenburg2001ontological} or concept vectors \cite{kastrati2019performance}. Another aspect regarding the future work is multilingual compatibility \cite{ghafoor2021impact}. TextBlob classifies non-English languages by using Google translate. This work to some extent, but can affect the validity of the original resource \cite{Kia2016Multilingual}. In that case, an improvement of the proposed tool will be to include sentiment algorithms which supports different languages. Lastly, the OSN analysis dashboard only covers Twitter data. To be able to understand peoples opinions about certain topics, it is required to include other social media platforms such as Instagram, Facebook and Reddit.

\bibliography{access.bib}{}

\begin{thebibliography}{10}
\providecommand{\url}[1]{#1}
\csname url@samestyle\endcsname
\providecommand{\newblock}{\relax}
\providecommand{\bibinfo}[2]{#2}
\providecommand{\BIBentrySTDinterwordspacing}{\spaceskip=0pt\relax}
\providecommand{\BIBentryALTinterwordstretchfactor}{4}
\providecommand{\BIBentryALTinterwordspacing}{\spaceskip=\fontdimen2\font plus
\BIBentryALTinterwordstretchfactor\fontdimen3\font minus
  \fontdimen4\font\relax}
\providecommand{\BIBforeignlanguage}[2]{{%
\expandafter\ifx\csname l@#1\endcsname\relax
\typeout{** WARNING: IEEEtran.bst: No hyphenation pattern has been}%
\typeout{** loaded for the language `#1'. Using the pattern for}%
\typeout{** the default language instead.}%
\else
\language=\csname l@#1\endcsname
\fi
#2}}
\providecommand{\BIBdecl}{\relax}
\BIBdecl

\bibitem{fang2015sentiment}
X.~Fang and J.~Zhan, ``Sentiment analysis using product review data,''
  \emph{Journal of Big Data}, vol.~2, no.~1, pp. 1--14, 2015.

\bibitem{bibi2022novel}
M.~Bibi, W.~A. Abbasi, W.~Aziz, S.~Khalil, M.~Uddin, C.~Iwendi, and T.~R.
  Gadekallu, ``A novel unsupervised ensemble framework using concept-based
  linguistic methods and machine learning for twitter sentiment analysis,''
  \emph{Pattern Recognition Letters}, vol. 158, pp. 80--86, 2022.

\bibitem{zhao2017weakly}
W.~Zhao, Z.~Guan, L.~Chen, X.~He, D.~Cai, B.~Wang, and Q.~Wang,
  ``Weakly-supervised deep embedding for product review sentiment analysis,''
  \emph{IEEE Transactions on Knowledge and Data Engineering}, vol.~30, no.~1,
  pp. 185--197, 2017.

\bibitem{10.1145}
Z.~Kastrati, B.~Arifaj, A.~Lubishtani, F.~Gashi, and E.~Nishliu, ``Aspect-based
  opinion mining of students' reviews on online courses,'' in \emph{Proceedings
  of the 2020 6th International Conference on Computing and Artificial
  Intelligence}.\hskip 1em plus 0.5em minus 0.4em\relax New York, NY, USA: ACM,
  2020, p. 510–514.

\bibitem{sadriu2022automated}
S.~Sadriu, K.~P. Nuci, A.~S. Imran, I.~Uddin, and M.~Sajjad, ``An automated
  approach for analysing students feedback using sentiment analysis
  techniques,'' in \emph{Mediterranean Conference on Pattern Recognition and
  Artificial Intelligence}.\hskip 1em plus 0.5em minus 0.4em\relax Springer,
  2022, pp. 228--239.

\bibitem{edalati2021potential}
M.~Edalati, A.~S. Imran, Z.~Kastrati, and S.~M. Daudpota, ``The potential of
  machine learning algorithms for sentiment classification of students’
  feedback on {MOOC},'' in \emph{Proceedings of SAI Intelligent Systems
  Conference}.\hskip 1em plus 0.5em minus 0.4em\relax Springer, 2021, pp.
  11--22.

\bibitem{feldman2011stock}
R.~Feldman, B.~Rosenfeld, R.~Bar-Haim, and M.~Fresko, ``The stock
  sonar—sentiment analysis of stocks based on a hybrid approach,'' in
  \emph{Proceedings of the AAAI Conference on Artificial Intelligence},
  vol.~25, no.~2, 2011, pp. 1642--1647.

\bibitem{batra2021evaluating}
R.~Batra, A.~S. Imran, Z.~Kastrati, A.~Ghafoor, S.~M. Daudpota, and S.~Shaikh,
  ``Evaluating polarity trend amidst the coronavirus crisis in peoples’
  attitudes toward the vaccination drive,'' \emph{Sustainability}, vol.~13,
  no.~10, p. 5344, 2021.

\bibitem{kastrati2021sentiment}
Z.~Kastrati, F.~Dalipi, A.~S. Imran, K.~Pireva~Nuci, and M.~A. Wani,
  ``Sentiment analysis of students’ feedback with {NLP} and deep learning: A
  systematic mapping study,'' \emph{Applied Sciences}, vol.~11, no.~9, p. 3986,
  2021.

\bibitem{sandra2022university}
L.~Sandra, G.~Gunarso, O.~W. Riruma \emph{et~al.}, ``Are university students
  independent: Twitter sentiment analysis of independent learning in
  independent campus using roberta base indolem sentiment classifier model,''
  in \emph{2021 International Seminar on Machine Learning, Optimization, and
  Data Science (ISMODE)}.\hskip 1em plus 0.5em minus 0.4em\relax IEEE, 2022,
  pp. 249--253.

\bibitem{kastrati2020weakly}
Z.~Kastrati, A.~S. Imran, and A.~Kurti, ``Weakly supervised framework for
  aspect-based sentiment analysis on students’ reviews of {MOOCs},''
  \emph{IEEE Access}, vol.~8, pp. 106\,799--106\,810, 2020.

\bibitem{nezhad2022twitter}
Z.~B. Nezhad and M.~A. Deihimi, ``Twitter sentiment analysis from iran about
  covid 19 vaccine,'' \emph{Diabetes \& Metabolic Syndrome: Clinical Research
  \& Reviews}, vol.~16, no.~1, p. 102367, 2022.

\bibitem{shariq2020cross}
A.~S. Imran, S.~M. Daudpota, Z.~Kastrati, and Batra, ``Cross-cultural polarity
  and emotion detection using sentiment analysis and deep learning--a case
  study on covid-19,'' \emph{arXiv e-prints}, pp. arXiv--2008, 2020.

\bibitem{ali2019sentiment}
N.~M. Ali, M.~M. Abd El~Hamid, and A.~Youssif, ``Sentiment analysis for movies
  reviews dataset using deep learning models,'' \emph{International Journal of
  Data Mining \& Knowledge Management Process (IJDKP) Vol}, vol.~9, 2019.

\bibitem{yasen2019movies}
M.~Yasen and S.~Tedmori, ``Movies reviews sentiment analysis and
  classification,'' in \emph{2019 IEEE Jordan International Joint Conference on
  Electrical Engineering and Information Technology (JEEIT)}.\hskip 1em plus
  0.5em minus 0.4em\relax IEEE, 2019, pp. 860--865.

\bibitem{Fu:2016}
K.-W. Fu, H.~Liang, N.~Saroha, Z.~T.~H. Tse, P.~Ip, and I.~C.-H. Fung, ``How
  people react to zika virus outbreaks on twitter? a computational content
  analysis,'' \emph{\textit{American Journal of Infection Control}}, vol.~44,
  no.~12, pp. 1700--1702, 2016.

\bibitem{imran2020cross}
A.~S. Imran, S.~M. Daudpota, Z.~Kastrati, and R.~Batra, ``Cross-cultural
  polarity and emotion detection using sentiment analysis and deep learning on
  covid-19 related tweets,'' \emph{Ieee Access}, vol.~8, pp.
  181\,074--181\,090, 2020.

\bibitem{electronics10101133}
Z.~Kastrati, L.~Ahmedi, A.~Kurti, F.~Kadriu, D.~Murtezaj, and F.~Gashi, ``A
  deep learning sentiment analyser for social media comments in low-resource
  languages,'' \emph{Electronics}, vol.~10, no.~10, 2021.

\bibitem{stamm2022does}
B.~Stamm and K.~Loomis, ``What does it mean to “engage” for learning on
  social media?: an analysis of global read aloud twitter participation,''
  \emph{Journal of Research on Technology in Education}, pp. 1--18, 2022.

\bibitem{liu_2015}
B.~Liu, \emph{Frontmatter}.\hskip 1em plus 0.5em minus 0.4em\relax Cambridge
  University Press, 2015.

\bibitem{dalipi2017analysis}
F.~Dalipi, A.~S. Imran, F.~Idrizi, and H.~Aliu, ``An analysis of learner
  experience with {MOOCs} in mobile and desktop learning environment,'' in
  \emph{Advances in human factors, business management, training and
  education}.\hskip 1em plus 0.5em minus 0.4em\relax Springer, 2017, pp.
  393--402.

\bibitem{JohnsonLaird2006AreTC}
P.~N. Johnson-Laird and N.~Y.~L. Lee, ``Are there cross-cultural differences in
  reasoning?'' 2006.

\bibitem{doi:10.1177/1094670509344213}
K.~Pauwels, T.~Ambler, B.~H. Clark, P.~LaPointe, D.~Reibstein, B.~Skiera,
  B.~Wierenga, and T.~Wiesel, ``Dashboards as a service: Why, what, how, and
  what research is needed?'' \emph{Journal of Service Research}, vol.~12,
  no.~2, pp. 175--189, 2009.

\bibitem{AzureGroup2020}
\BIBentryALTinterwordspacing
A.~Group. (2020) The importance of dashboards in business: Why use dashboard
  reports? [Online]. Available:
  \url{https://blog.azuregroup.com.au/the-importance-of-dashboard-reporting-in-business-why-use-dashboard-reports}
\BIBentrySTDinterwordspacing

\bibitem{NielsenNormanGroup2020}
\BIBentryALTinterwordspacing
N.~N. Group. (2020) Data visualizations for dashboards. [Online]. Available:
  \url{https://www.nngroup.com/videos/data-visualizations-dashboards/,}
\BIBentrySTDinterwordspacing

\bibitem{BrianDean2022}
\BIBentryALTinterwordspacing
B.~Dean. (2022) How many people use twitter in 2022. [Online]. Available:
  \url{https://backlinko.com/twitter-users\#twitter-users, January 2022.}
\BIBentrySTDinterwordspacing

\bibitem{9237640}
J.~Yang and J.~Yang, ``Aspect based sentiment analysis with self-attention and
  gated convolutional networks,'' in \emph{2020 IEEE 11th International
  Conference on Software Engineering and Service Science (ICSESS)}, 2020, pp.
  146--149.

\bibitem{8806182}
J.~Kim, J.~Seo, M.~Lee, and J.~Seok, ``Stock price prediction through the
  sentimental analysis of news articles,'' in \emph{2019 Eleventh International
  Conference on Ubiquitous and Future Networks (ICUFN)}, 2019, pp. 700--702.

\bibitem{8621884}
D.~Shah, H.~Isah, and F.~Zulkernine, ``Predicting the effects of news
  sentiments on the stock market,'' in \emph{2018 IEEE International Conference
  on Big Data (Big Data)}, 2018, pp. 4705--4708.

\bibitem{8537242}
T.~Mankar, T.~Hotchandani, M.~Madhwani, A.~Chidrawar, and C.~Lifna, ``Stock
  market prediction based on social sentiments using machine learning,'' in
  \emph{2018 International Conference on Smart City and Emerging Technology
  (ICSCET)}, 2018, pp. 1--3.

\bibitem{8399007}
F.~Nausheen and S.~H. Begum, ``Sentiment analysis to predict election results
  using python,'' in \emph{2018 2nd International Conference on Inventive
  Systems and Control (ICISC)}, 2018, pp. 1259--1262.

\bibitem{Maite2011Lexicon}
M.~T. K. V. M.~S. Maite~Taboada, Julian~Brooke, ``Lexicon-based methods for
  sentiment analysis.'' \emph{Computational Linguistics}, vol.~37, pp.
  267--307, 2011.

\bibitem{9277094}
S.~Yadav and N.~Saleena, ``Sentiment analysis of reviews using an augmented
  dictionary approach,'' in \emph{2020 5th International Conference on
  Computing, Communication and Security (ICCCS)}, 2020, pp. 1--5.

\bibitem{Zhang2018Deep}
L.~Zhang, S.~Wang, and B.~Liu, ``Deep learning for sentiment analysis : A
  survey,'' 2018.

\bibitem{DBLP:journals/corr/abs-1709-02984}
F.~Calefato, F.~Lanubile, F.~Maiorano, and N.~Novielli, ``Sentiment polarity
  detection for software development,'' \emph{CoRR}, vol. abs/1709.02984, 2017.

\bibitem{rajput2016lexicon}
Q.~Rajput, S.~Haider, and S.~Ghani, ``Lexicon-based sentiment analysis of
  teachers’ evaluation,'' \emph{Applied computational intelligence and soft
  computing}, vol. 2016, 2016.

\bibitem{puigcerver2017querying}
J.~Puigcerver, A.~H. Toselli, and E.~Vidal, ``Querying out-of-vocabulary words
  in lexicon-based keyword spotting,'' \emph{Neural Computing and
  Applications}, vol.~28, no.~9, pp. 2373--2382, 2017.

\bibitem{kastrati2016semcon}
Z.~Kastrati, A.~S. Imran, and S.~Yildirim-Yayilgan, ``{SEMCON}: a semantic and
  contextual objective metric for enriching domain ontology concepts,''
  \emph{International Journal on Semantic Web and Information Systems
  (IJSWIS)}, vol.~12, no.~2, pp. 1--24, 2016.

\bibitem{tsatsaronis2010semanticrank}
G.~Tsatsaronis, I.~Varlamis, and K.~N{\o}rv{\aa}g, ``Semanticrank: ranking
  keywords and sentences using semantic graphs,'' in \emph{Proceedings of the
  23rd international conference on computational linguistics (Coling 2010)},
  2010, pp. 1074--1082.

\bibitem{kastrati2015semcon}
Z.~Kastrati, A.~S. Imran, and S.~Y. Yayilgan, ``{SEMCON}: semantic and
  contextual objective metric,'' in \emph{Proceedings of the 2015 IEEE 9th
  International Conference on Semantic Computing (IEEE ICSC 2015)}.\hskip 1em
  plus 0.5em minus 0.4em\relax IEEE, 2015, pp. 65--68.

\bibitem{chandio2022attention}
B.~A. Chandio, A.~S. Imran, M.~Bakhtyar, S.~M. Daudpota, and J.~Baber,
  ``Attention-based ru-bilstm sentiment analysis model for roman urdu,''
  \emph{Applied Sciences}, vol.~12, no.~7, p. 3641, 2022.

\bibitem{batra2021large}
R.~Batra, Z.~Kastrati, A.~S. Imran, S.~M. Daudpota, and A.~Ghafoor, ``A
  large-scale tweet dataset for urdu text sentiment analysis,'' \emph{arXiv
  e-prints}, pp. arXiv--2021, 2021.

\bibitem{10.1007/978-1-4614-6880-6_28}
S.~Zhu, B.~Xu, D.~Zheng, and T.~Zhao, ``Chinese microblog sentiment analysis
  based on semi-supervised learning,'' in \emph{Semantic Web and Web Science},
  J.~Li, G.~Qi, D.~Zhao, W.~Nejdl, and H.-T. Zheng, Eds.\hskip 1em plus 0.5em
  minus 0.4em\relax New York, NY: Springer New York, 2013, pp. 325--331.

\bibitem{fatima2022systematic}
N.~Fatima, A.~S. Imran, Z.~Kastrati, S.~M. Daudpota, A.~Soomro, and S.~Shaikh,
  ``A systematic literature review on text generation using deep neural network
  models,'' \emph{IEEE Access}, 2022.

\bibitem{Kia2016Multilingual}
A.~H. E. C. A. Y. A. H. A. G. . Q.~Z. Kia~Dashtipour, Soujanya~Poria,
  ``Multilingual sentiment analysis: State of the art and independent
  comparison of techniques,'' \emph{Cognitive Computation}, vol.~7, pp.
  757--771, 2016.

\bibitem{ghafoor2021impact}
A.~Ghafoor, A.~S. Imran, S.~M. Daudpota, Z.~Kastrati, R.~Batra, M.~A. Wani
  \emph{et~al.}, ``The impact of translating resource-rich datasets to
  low-resource languages through multi-lingual text processing,'' \emph{IEEE
  Access}, vol.~9, pp. 124\,478--124\,490, 2021.

\bibitem{imran2022impact}
A.~S. Imran, R.~Yang, Z.~Kastrati, S.~M. Daudpota, and S.~Shaikh, ``The impact
  of synthetic text generation for sentiment analysis using gan based models,''
  \emph{Egyptian Informatics Journal}, 2022.

\bibitem{8628718}
D.~Vilares, H.~Peng, R.~Satapathy, and E.~Cambria, ``Babelsenticnet: A
  commonsense reasoning framework for multilingual sentiment analysis,'' in
  \emph{2018 IEEE Symposium Series on Computational Intelligence (SSCI)}, 2018,
  pp. 1292--1298.

\bibitem{ijtech-2753}
M.~N. M.~N. Nurul Husna~Mahadzir, Mohd Faizal~Omar, ``A sentiment analysis
  visualization system for the property industry,'' \emph{International Journal
  of Technology}, vol.~9, no.~8, pp. 1609--1617, 2018.

\bibitem{7919584}
K.~L.~S. Kumar, J.~Desai, and J.~Majumdar, ``Opinion mining and sentiment
  analysis on online customer review,'' in \emph{2016 IEEE International
  Conference on Computational Intelligence and Computing Research (ICCIC)},
  2016, pp. 1--4.

\bibitem{Google2021}
\BIBentryALTinterwordspacing
Google. (2021) Introduction to microservices — google cloud. [Online].
  Available:
  \url{https://cloud.google.com/architecture/microservices-architecture-introduction}
\BIBentrySTDinterwordspacing

\bibitem{10635_99059}
S.~Sasidharan and G.~S. Dhanesh, ``The role of color in influencing trust the
  role of color in influencing trust in e-commerce web site,'' in
  \emph{"Proceedings of the Second Midwest United States Association for
  Information Systems, Springfield, IL May 18–19, 2007."}, 2007.

\bibitem{Softqube2022}
\BIBentryALTinterwordspacing
Softqube. (2022) New innovative approach of problem solving. [Online].
  Available: \url{https://www.softqubes.com/blog/triz}
\BIBentrySTDinterwordspacing

\bibitem{Council2020}
\BIBentryALTinterwordspacing
T.~R.~C. of~Norway. (2020) Scale of marks and assessment criteria. [Online].
  Available:
  \url{https://www.forskningsradet.no/en/processing-grant-applications/processing-applications/Scale-marks-assessment-criteria/}
\BIBentrySTDinterwordspacing

\bibitem{imran2012multimedia}
A.~S. Imran and F.~A. Cheikh, ``Multimedia learning objects framework for
  e-learning,'' in \emph{2012 International Conference on E-Learning and
  E-Technologies in Education (ICEEE)}.\hskip 1em plus 0.5em minus 0.4em\relax
  IEEE, 2012, pp. 105--109.

\bibitem{misuraca2020sentiment}
M.~Misuraca, A.~Forciniti, G.~Scepi, and M.~Spano, ``Sentiment analysis for
  education with r: packages, methods and practical applications,'' \emph{arXiv
  preprint arXiv:2005.12840}, 2020.

\bibitem{imran2014hip}
A.~S. Imran and S.~J. Kowalski, ``{HIP}--a technology-rich and interactive
  multimedia pedagogical platform,'' in \emph{International Conference on
  Learning and Collaboration Technologies}.\hskip 1em plus 0.5em minus
  0.4em\relax Springer, 2014, pp. 151--160.

\bibitem{COSTOLA2021110021}
M.~Costola, M.~Iacopini, and C.~R. Santagiustina, ``On the “mementum” of
  meme stocks,'' \emph{Economics Letters}, vol. 207, p. 110021, 2021.

\bibitem{kastrati2020wet}
Z.~Kastrati, A.~Kurti, and A.~S. Imran, ``{WET}: Word embedding-topic
  distribution vectors for {MOOC} video lectures dataset,'' \emph{Data in
  brief}, vol.~28, p. 105090, 2020.

\bibitem{zhang2018deep12}
L.~Zhang, S.~Wang, and B.~Liu, ``Deep learning for sentiment analysis: A
  survey,'' \emph{Wiley Interdisciplinary Reviews: Data Mining and Knowledge
  Discovery}, vol.~8, no.~4, p. e1253, 2018.

\bibitem{sun2019utilizing}
C.~Sun, L.~Huang, and X.~Qiu, ``Utilizing bert for aspect-based sentiment
  analysis via constructing auxiliary sentence,'' \emph{arXiv preprint
  arXiv:1903.09588}, 2019.

\bibitem{kastrati2015improved}
Z.~Kastrati, A.~S. Imran, and S.~Y. Yayilgan, ``An improved concept vector
  space model for ontology based classification,'' in \emph{2015 11th
  International Conference on Signal-Image Technology \& Internet-Based Systems
  (SITIS)}.\hskip 1em plus 0.5em minus 0.4em\relax IEEE, 2015, pp. 240--245.

\bibitem{nirenburg2001ontological}
S.~Nirenburg and V.~Raskin, ``Ontological semantics, formal ontology, and
  ambiguity,'' in \emph{Proceedings of the international conference on Formal
  Ontology in Information Systems-Volume 2001}, 2001, pp. 151--161.

\bibitem{kastrati2019performance}
Z.~Kastrati and A.~S. Imran, ``Performance analysis of machine learning
  classifiers on improved concept vector space models,'' \emph{Future
  Generation Computer Systems}, vol.~96, pp. 552--562, 2019.

\end{thebibliography}


\begin{thebibliography}{10}
\providecommand{\url}[1]{#1}
\csname url@samestyle\endcsname
\providecommand{\newblock}{\relax}
\providecommand{\bibinfo}[2]{#2}
\providecommand{\BIBentrySTDinterwordspacing}{\spaceskip=0pt\relax}
\providecommand{\BIBentryALTinterwordstretchfactor}{4}
\providecommand{\BIBentryALTinterwordspacing}{\spaceskip=\fontdimen2\font plus
\BIBentryALTinterwordstretchfactor\fontdimen3\font minus
  \fontdimen4\font\relax}
\providecommand{\BIBforeignlanguage}[2]{{%
\expandafter\ifx\csname l@#1\endcsname\relax
\typeout{** WARNING: IEEEtran.bst: No hyphenation pattern has been}%
\typeout{** loaded for the language `#1'. Using the pattern for}%
\typeout{** the default language instead.}%
\else
\language=\csname l@#1\endcsname
\fi
#2}}
\providecommand{\BIBdecl}{\relax}
\BIBdecl

\bibitem{johnson2006there}
P.~Johnson-Laird and N.~Lee, ``Are there cross-cultural differences in
  reasoning?'' in \emph{\textit{Proceedings of the Annual Meeting of the
  Cognitive Science Society}}, 2006, pp. 459--464.

\bibitem{nisbett2004geography}
R.~Nisbett, \emph{\textit{The geography of thought: How Asians and Westerners
  think differently... and why}}.\hskip 1em plus 0.5em minus 0.4em\relax Simon
  and Schuster, 2004.

\bibitem{Local:2020}
\BIBentryALTinterwordspacing
T.~L. Dk, ``Why is denmark's coronavirus lockdown so much tougher than
  sweden's?'' \emph{\textit{The Local}}. [Online]. Available:
  \url{https://www.thelocal.dk/20200320/why-is-denmarks-lockdown-so-much-more-severe-than-swedens}
\BIBentrySTDinterwordspacing

\bibitem{hinton2006fast}
G.~E. Hinton, S.~Osindero, and Y.-W. Teh, ``A fast learning algorithm for deep
  belief nets,'' \emph{\textit{Neural computation}}, vol.~18, no.~7, pp.
  1527--1554, 2006.

\bibitem{Dunkel:2019}
A.~Dunkel, G.~Andrienko, N.~Andrienko, D.~Burghardt, E.~Hauthal, and R.~Purves,
  ``A conceptual framework for studying collective reactions to events in
  location-based social media,'' \emph{\textit{International Journal of
  Geographical Information Science}}, vol.~33, no.~4, pp. 780--804, 2019.

\bibitem{Liang:2019}
H.~Liang, I.~C.-H. Fung, Z.~T.~H. Tse, J.~Yin, C.-H. Chan, L.~E. Pechta, B.~J.
  Smith, R.~D. Marquez-Lamed, M.~I. Meltzer, K.~M. Lubell, and K.-W. Fu, ``How
  did ebola information spread on twitter: broadcasting or viral spreading?''
  \emph{\textit{BMC Public Health}}, vol.~19, no. 438, pp. 1--11, 2019.

\bibitem{Kaila:2020}
R.~P. Kaila and A.~K. Prasad, ``Informational flow on twitter – corona virus
  outbreak – topic modelling approach,'' \emph{\textit{International Journal
  of Advanced Research in Engineering and Technology (IJARET)}}, vol.~11,
  no.~3, pp. 128--134, 2020.

\bibitem{Szomszor:2011}
M.~{Szomszor}, P.~{Kostkova}, and C.~S. {Louis}, ``Twitter informatics:
  Tracking and understanding public reaction during the 2009 swine flu
  pandemic,'' in \emph{\textit{Proceedings of the IEEE/WIC/ACM International
  Conferences on Web Intelligence and Intelligent Agent Technology}}, vol.~1,
  2011, pp. 320--323.

\bibitem{Fu:2016}
K.-W. Fu, H.~Liang, N.~Saroha, Z.~T.~H. Tse, P.~Ip, and I.~C.-H. Fung, ``How
  people react to zika virus outbreaks on twitter? a computational content
  analysis,'' \emph{\textit{American Journal of Infection Control}}, vol.~44,
  no.~12, pp. 1700--1702, 2016.

\bibitem{Vorovchenko:2017}
T.~Vorovchenko, P.~Ariana, F.~van Loggerenberg, and P.~Amirian, ``{\#}ebola and
  twitter. what insights can global health draw from social media?'' in
  \emph{\textit{Proceedings of the Big Data in Healthcare: Extracting Knowledge
  from Point-of-Care Machines}}, P.~Amirian, T.~Lang, and F.~van Loggerenberg,
  Eds., 2017, pp. 85--98.

\bibitem{Chew:2009}
C.~Chew and G.~Eysenbach, ``Pandemics in the age of twitter: Content analysis
  of tweets during the 2009 h1n1 outbreak,'' \emph{\textit{PLoS ONE}}, vol.~5,
  no.~11, pp. 1--13, 2010.

\bibitem{shelar2018sentiment}
A.~Shelar and C.-Y. Huang, ``Sentiment analysis of twitter data,'' in
  \emph{\textit{Proceedings of the 2018 International Conference on
  Computational Science and Computational Intelligence (CSCI)}}, 2018, pp.
  1301--1302.

\bibitem{9110884}
Z.~{Kastrati}, A.~S. {Imran}, and A.~{Kurti}, ``Weakly supervised framework for
  aspect-based sentiment analysis on students’ reviews of moocs,''
  \emph{\textit{IEEE Access}}, vol.~8, pp. 106\,799--106\,810, 2020.

\bibitem{das2018real}
S.~Das, R.~K. Behera, S.~K. Rath \emph{et~al.}, ``Real-time sentiment analysis
  of twitter streaming data for stock prediction,'' \emph{\textit{Procedia
  computer science}}, vol. 132, pp. 956--964, 2018.

\bibitem{pagolu2016sentiment}
V.~S. Pagolu, K.~N. Reddy, G.~Panda, and B.~Majhi, ``Sentiment analysis of
  twitter data for predicting stock market movements,'' in
  \emph{\textit{Proceedings of the international conference on signal
  processing, communication, power and embedded system (SCOPES)}}, 2016, pp.
  1345--1350.

\bibitem{batra2018integrating}
R.~Batra and S.~M. Daudpota, ``Integrating stocktwits with sentiment analysis
  for better prediction of stock price movement,'' in \emph{\textit{Proceedings
  of the 2018 International Conference on Computing, Mathematics and
  Engineering Technologies (iCoMET)}}, 2018, pp. 1--5.

\bibitem{budiharto2018prediction}
W.~Budiharto and M.~Meiliana, ``Prediction and analysis of indonesia
  presidential election from twitter using sentiment analysis,''
  \emph{\textit{Journal of Big data}}, vol.~5, no.~1, p.~51, 2018.

\bibitem{liao2017cnn}
S.~Liao, J.~Wang, R.~Yu, K.~Sato, and Z.~Cheng, ``Cnn for situations
  understanding based on sentiment analysis of twitter data,''
  \emph{\textit{Procedia computer science}}, vol. 111, pp. 376--381, 2017.

\bibitem{musto2014comparison}
C.~Musto, G.~Semeraro, and M.~Polignano, ``A comparison of lexicon-based
  approaches for sentiment analysis of microblog posts,'' in
  \emph{\textit{Proceedings of the 8th International Workshop on Information
  Filtering and Retrieval}}, 2014, pp. 59--68.

\bibitem{kharde2016sentiment}
V.~Kharde and S.~Sonawane, ``Sentiment analysis of twitter data: a survey of
  techniques,'' \emph{\textit{arXiv preprint arXiv:1601.06971}}, pp. 5--15,
  2016.

\bibitem{zhang2018deep}
L.~Zhang, S.~Wang, and B.~Liu, ``Deep learning for sentiment analysis: A
  survey,'' \emph{\textit{Wiley Interdisciplinary Reviews: Data Mining and
  Knowledge Discovery}}, vol.~8, no.~4, pp. 1--34, 2018.

\bibitem{giachanou2016like}
A.~Giachanou and F.~Crestani, ``Like it or not: A survey of twitter sentiment
  analysis methods,'' \emph{\textit{ACM Computing Surveys (CSUR)}}, vol.~49,
  no.~2, pp. 1--41, 2016.

\bibitem{desai2016techniques}
M.~Desai and M.~A. Mehta, ``Techniques for sentiment analysis of twitter data:
  A comprehensive survey,'' in \emph{\textit{Proceedings of the International
  Conference on Computing, Communication and Automation (ICCCA)}}, 2016, pp.
  149--154.

\bibitem{Xiang:2015}
X.~Ji, S.~A. Chun, Z.~Wei, and J.~Geller, ``Twitter sentiment classification
  for measuring public health concerns,'' \emph{\textit{Social Network Analysis
  and Mining}}, vol.~5, no.~1, pp. 1--25, 2015.

\bibitem{Samuel:2020}
J.~Samuel, M.~N. Ali, M.~M. Rahman, E.~Esawi, and Y.~Samuel, ``Covid-19 public
  sentiment insights and machine learning for tweets classification,''
  \emph{\textit{Information}}, vol.~11, pp. 1--21, 2020.

\bibitem{Barkur:2020}
G.~Barkur, Vibha, and G.~B. Kamath, ``Sentiment analysis of nationwide lockdown
  due to covid 19 outbreak: Evidence from india,'' \emph{\textit{Asian Journal
  of Psychiatry}}, vol.~51, pp. 1--2, 2020.

\bibitem{hasan2014emotex}
M.~Hasan, E.~Rundensteiner, and E.~Agu, ``Emotex: Detecting emotions in twitter
  messages,'' in \emph{\textit{Proceedings of the ASE
  Bigdata/Socialcom/Cybersecurity Conference}}, 2014, pp. 1--10.

\bibitem{Fung:2014}
I.~C.-H. Fung, Z.~T.~H. Tse, C.-N. Cheung, A.~S. Miu, and K.-W. Fu, ``Ebola and
  the social media,'' \emph{\textit{The Lancet}}, vol. 384, no. 9961, pp.
  128--134, 2014.

\bibitem{do2016analyzing}
H.~J. Do, C.-G. Lim, Y.~J. Kim, and H.-J. Choi, ``Analyzing emotions in twitter
  during a crisis: A case study of the 2015 middle east respiratory syndrome
  outbreak in korea,'' in \emph{\textit{Proceedings of the international
  conference on big data and smart computing (BigComp)}}, 2016, pp. 415--418.

\bibitem{go2009twitter}
A.~Go, R.~Bhayani, and L.~Huang, ``\textit{Twitter sentiment classification
  using distant supervision},'' \emph{CS224N project report, Stanford}, vol.~1,
  no.~12, pp. 1--6, 2009.

\bibitem{MohammadB17wassa}
S.~M. Mohammad and F.~Bravo-Marquez, ``{WASSA-2017} shared task on emotion
  intensity,'' in \emph{\textit{Proceedings of the Workshop on Computational
  Approaches to Subjectivity, Sentiment and Social Media Analysis (WASSA)}},
  2017, pp. 34--39.

\bibitem{caisentiment}
M.~Cai, ``Sentiment analysis of tweets using deep neural architectures,'' in
  \emph{\textit{Proceedings of the 32nd Conference on Neural Information
  Processing Systems (NIPS 2018)}}, 2018, pp. 1--8.

\bibitem{pennington2014glove}
J.~Pennington, R.~Socher, and C.~D. Manning, ``Glove: Global vectors for word
  representation,'' in \emph{\textit{Proceedings of the 2014 conference on
  empirical methods in natural language processing (EMNLP)}}, 2014, pp.
  1532--1543.

\bibitem{Kastrati:2019ptr}
Z.~Kastrati, A.~S. Imran, and A.~Kurti, ``Integrating word embeddings and
  document topics with deep learning in a video classification framework,''
  \emph{\textit{Pattern Recognition Letters}}, vol. 128, pp. 85 -- 92, 2019.

\bibitem{debnath2020semantic}
A.~Debnath, N.~Pinnaparaju, M.~Shrivastava, V.~Varma, and I.~Augenstein,
  ``Semantic textual similarity of sentences with emojis,'' in
  \emph{\textit{Companion Proceedings of the Web Conference 2020}}, 2020, pp.
  426--430.

\bibitem{chang2020impact}
W.-L. Chang and H.-C. Tseng, ``The impact of sentiment on content post
  popularity through emoji and text on social platforms,'' in
  \emph{\textit{Proceedings of the Cyber Influence and Cognitive Threats}},
  2020, pp. 159--184.

\bibitem{Kastrati:2015SITIS}
Z.~{Kastrati}, A.~S. {Imran}, and S.~Y. {Yayilgan}, ``An improved concept
  vector space model for ontology based classification,'' in \emph{2015 11th
  International Conference on Signal-Image Technology Internet-Based Systems
  (SITIS)}, 2015, pp. 240--245.

\bibitem{kastrati2019impact}
Z.~Kastrati, A.~S. Imran, and S.~Y. Yayilgan, ``The impact of deep learning on
  document classification using semantically rich representations,''
  \emph{\textit{Information Processing \& Management}}, vol.~56, no.~5, pp.
  1618--1632, 2019.

\end{thebibliography}
\bibliographystyle{IEEEtran}

\end{document}